\pdfoutput=1
\documentclass[11pt]{article}

\usepackage[preprint]{acl}

\usepackage{times}
\usepackage{latexsym}

\usepackage[T1]{fontenc}
\usepackage[utf8]{inputenc}

\usepackage{microtype}

\usepackage{inconsolata}

\usepackage{comment}
\usepackage{adjustbox}
\usepackage{multirow}
\usepackage{booktabs}
\usepackage{csquotes}
\usepackage{amsmath}
\usepackage{graphicx}
\graphicspath{ {./images/} }
\usepackage{svg}
\usepackage{times}
\usepackage{latexsym}
\usepackage{natbib}
\usepackage{threeparttable}
\usepackage{xcolor}
\usepackage{microtype}

\title{Instruction Finetuning for Leaderboard Generation \\ from Empirical AI Research}

\author{Salomon Kabongo \\
  Leibniz University of Hannover \\
  Hannover, Germany \\
  \texttt{kabenamualu@l3s.de} \\\And
  Jennifer D'Souza \\
  TIB\\
  Hannover, Germany \\
  \texttt{jennifer.dsouza@tib.eu} \\}

\begin{document}
\maketitle
\begin{abstract}
This study demonstrates the application of instruction finetuning of pretrained Large Language Models (LLMs) to automate the generation of AI research leaderboards, extracting (Task, Dataset, Metric, Score) quadruples from articles. It aims to streamline the dissemination of advancements in AI research by transitioning from traditional, manual community curation, or otherwise taxonomy-constrained natural language inference (NLI) models, to an automated, generative LLM-based approach. Utilizing the FLAN-T5 model, this research enhances LLMs' adaptability and reliability in information extraction, offering a novel method for structured knowledge representation.
\end{abstract}

\section{Introduction}

The burgeoning complexity and volume of scientific literature~\cite{fortunato2018science,bornmann2021growth,altbach2019too} necessitate sophisticated methods for distilling and structuring vast amounts of data~\cite{auer2020improving}, particularly in fields like Artificial Intelligence (AI) research. Instruction finetuning of Large Language Models (LLMs) emerges as a pivotal innovation, addressing this need by honing models' abilities to precisely interpret~\cite{wei2021finetuned} and execute specific instructions for tasks such as information extraction. This precision is not just a technical requirement but a transformative approach to how models interact with and process the unstructured text, shifting the paradigm from broad, conversational responses to targeted, information-rich outputs. Recent studies~\cite{lu-etal-2023-pivoine,wang2023instructuie} underscore the importance of fine-tuning in guiding LLMs to better understand and respond to nuanced task-specific directives, thereby enhancing their utility across diverse research and industry applications.

%See \autoref{fig:tdms-example}.

At the heart of this study is the State-Of-The-Art (SOTA) task, an innovative venture into extracting leaderboards from empirical AI research publications in the form of (Task, Dataset, Metric, Score) quadruples, or (T, D, M, S) henceforward~\cite{hou-etal-2019-identification}. Leaderboards serve as a critical tool for benchmarking and navigating scientific progress. Traditional leaderboards have been community-curated, exemplified by platforms like \href{https://paperswithcode.com/}{PapersWithCode} (PwC) or Open Research Knowledge Graph's \href{https://orkg.org/benchmarks}{benchmarks} feature. However, text mining can expedite leaderboard construction, capturing the (T, D, M, S) quadruple information buried within the discourse of scholarly AI articles. Only two prior works, IBM-TDMS~\cite{hou-etal-2019-identification} and AxCell~\cite{axcell}, have assessed automated text mining systems for the (T, D, M, S) quadruple extraction task. IBM-TDMS achieved 7.5 micro F1 and 8.8 macro F1 scores, while AxCell improved upon this with 25.8 micro F1 and 19.7 macro F1. These systems treated (T, D, M, S) extraction as a Natural Language Inference (NLI) task, reliant on a predefined (T, D, M) taxonomy. The drawback of this approach is its inability to detect newly introduced (T, D, M) elements outside the taxonomy, rendering the systems impractical. In this work, we introduce a novel objective: \textit{text generation within a given context}, aiming to overcome these limitations. Furthermore, this work adopts instruction fine-tuning to accomplish SOTA as a text generation task, and enhance the model's adaptability to the domain-specific nuances of AI research. \textsc{SOTA}, in our work, aims to achieve two core goals: first, to determine if an article reports a leaderboard, and second, to extract relevant (T, D, M, S) quadruples within a generation framework. This innovative approach overcomes previous limitations of NLI systems, enabling us to detect newly introduced (T, D, M) elements and rendering our approach practically feasible. The remaining research question we address in this work is the challenge to move the needle in terms of performance on \textsc{SOTA} such that the system is indeed reliable in a practical setting.

In this study, we harness the capabilities of the FLAN-T5 model~\cite{flan-t5}, an instruction-tuned variant from the T5 model class~\cite{t5}, boasting 780M parameters and sourced from Google's open-access repository on the \href{https://huggingface.co/docs/transformers/model_doc/flan-t5}{Transformers library}. There could have been one of two directions for this work: scaling the models or instruction fine-tuning of a moderate-sized LLM, i.e. with parameters in millions versus 1000x more in billions. We chose the latter. We believe that our choice makes model tuning more accessible within the research community while empirically proving to be nonetheless effective (experimental details in \autoref{eval}). For instruction-based finetuning, we use applicable instructions from the \textit{open-sourced instruction generalization efforts} introduced as the ``Flan 2022 Collection''~\citep{flan-collection}. Our approach differs from finetuning a pretrained LM as we instead finetune an instruction-tuned LM, enabling the model to effectively follow instructions it has been trained on and adapt to a new domain and task, without the need to handle variability in learning new instruction formats. This methodological choice not only enhances the model's performance but also promotes reproducibility and innovation in automated information extraction and knowledge representation within AI research.

Summarily, our contributions include: 1) A novel methodological approach that employs ``single-task instruction-finetuning'' with the FLAN-T5 model, to enhance its domain and task adaptation. Our \href{https://anonymous.4open.science/r/LLLM-Leaderboard-ESWC-FLAN-T5/README.md}{source code} is released. 2) A departure from traditional NLI methods towards an LLM-based system that utilizes moderate-sized models for greater practical application viability. 3) The introduction of a new corpus for experimental validation, promoting standardized comparisons in future SOTA task research. 4) Demonstrated improvements in task performance, with our model surpassing previous NLI-based systems by nearly 10\% in F1 scores, thereby validating the efficacy and feasibility of our approach.

\section{Related Work}

%\subsection{Scientific Information Extraction (IE)}

At the heart of \textsc{SOTA} is a scientific information extraction (IE) task. Different from most previous work on IE from scientific literature which concentrates mainly on the titles or abstract section or individual paragraphs~\citep{ftd,acl-rd-tec,scienceie,sciie,cl-titles,cs-ner}, our task needs to analyze the entire paper addressing document-level IE. Relatedly, other works that address document-level IE via extraction objectives that are similar to our (T, D, M, S) is the IBM-TDMS system~\citep{hou-etal-2019-identification}, AxCell~\citep{axcell}, SciREX~\cite{scirex} which addresses (Task, Dataset, Method, Metric) replacing Score by Method, the ORKG-TDM~\citep{kabongo2023orkg,kabongo2023zero} and SciNLPKG~\citep{mondal-etal-2021-end} which address only the (T, D, M) objective. While \citep{hou-etal-2019-identification} addressed the (T, D, M, S) objective, their experimental dataset was relatively small and LLMs were not the focus of their experiments. Nevertheless, they seminally introduced the DocTAET context feature as a shorter, focused representation of the full paper in which the task, dataset, metric, and scores are most likely to be mentioned. The TAET in DocTAET represents the Title, Abstracts, Experimental setup, and Tables including captions and headers of the paper extracted with the help of customized heuristic-based extraction parsers and supplied as context to the machine learning model. This context representation is also used in our work. Notably, we employ LLMs for leaderboard construction and adopt an open-world assumption for text generation, a first in this context, moving away from the closed-world models reliant on a fixed (T, D, M) taxonomy. Constraining the (T, D, M) taxonomy via a closed-world assumption does not reflect the real-world where new tasks or datasets are constantly being introduced. Thus the traditional reported NLI models are not generalizable compared to our generation approach.

\begin{table*}[h]
\begin{center}
\begin{threeparttable}
\begin{minipage}{\textwidth}
\begin{tabular*}{\textwidth}{@{\extracolsep{\fill}}l|cccc@{\extracolsep{\fill}}}
% \toprule%
\cmidrule{1-3}\cmidrule{3-5}%
& \multicolumn{2}{@{}c@{}}{\textbf{Our Corpus}} & \multicolumn{2}{@{}c@{}}{\textbf{Prior Work}} \\\cmidrule{2-3}\cmidrule{4-5}%
 & Train & Test - Zeroshot & Train & Test set \\
\midrule
Papers w/ leaderboards & 7,987 & 241 & 170  &  167  \\
Papers w/o leaderboards &  4,401 & 548 & - &  - \\
Total TDM-triples & 415,788 & 14,800 & 327  & 294 \\
Distinct TDM-triples & 11,998 & 1,267 & 78  & 78 \\
Distinct \textit{Tasks}       & 1,374 & 236 & 18  & 18 \\
Distinct \textit{Datasets}    & 4,816 & 647 & 44 & 44 \\
Distinct \textit{Metrics}     & 2,876 & 412 & 31 & 31 \\
Avg. no. of TDM per paper & 5.12 & 6.11 &  2.64 & 2.41 \\
Avg. no. of TDMS per paper & 6.95 & 7.86 &  - & - \\
% \botrule
\end{tabular*}
\caption{Our Corpus vs Prior Work~\cite{hou-etal-2019-identification} corpora statistics. The ``papers w/o leaderboard'' reffers to papers that do not report leaderboard .}
%\vspace{-20pt}
%\begin{tablenotes}
%    \item[a] Avg. number of TDM-triples per paper
%\end{tablenotes}
\label{table:dataset_stats}
\end{minipage}
\end{threeparttable}
\end{center}
\end{table*}

\section{Our Corpus}
\label{sec:corpus}

\noindent{\textbf{Corpus with (T, D, M, S) annotations.}} We created a new corpus as a collection of scholarly papers with their (T, D, M, S) quadruple annotations for evaluating the \textsc{SOTA} task~\cite{hou-etal-2019-identification}. This dataset is derived from the community-curated (T, D, M, S) annotations for thousands of AI articles available on PwC (CC BY-SA). Its articles span Natural Language Processing and Computer Vision domains, among other AI domains such as Robotics, Graphs, Reasoning, etc, thus, being representative for empirical AI research. The specific PwC source download timestamps is December 09, 2023. As such the corpus comprised over 7,500 articles. \noindent{\textbf{Corpus with (T, D, M, S) annotations.}} We created a new corpus as a collection of scholarly papers with their (T, D, M, S) quadruple annotations for evaluating the \textsc{SOTA} task~\cite{hou-etal-2019-identification}. This dataset is derived from the community-curated (T, D, M, S) annotations for thousands of AI articles available on PwC (CC BY-SA). Its articles span Natural Language Processing and Computer Vision domains, among other AI domains such as Robotics, Graphs, Reasoning, etc, thus, being representative for empirical AI research. The specific PwC source download timestamps is \textbf{December 09, 2023}. As such the corpus comprised over 7,500 articles. These articles, originally sourced from arXiv under CC-BY licenses, are available as latex code source, each accompanied by one or more (T, D, M, S) annotations from PwC. While the respective articles' metadata was directly obtained from the PwC data release, the articles collection had to be reconstructed by downloading them from arXiv under CC-BY licenses. Once downloaded, the articles being in .tex format needed to undergo pre-processing for tex-to-text conversion so that their contents could be mined. For this, the Pandoc alongside a custom script was applied to extract targeted regions of the paper DocTEAT which stands for \textbf{DOC}ument, \textbf{T}itle, \textbf{A}bstract, \textbf{E}xpSetup, and \textbf{T}ableInfo ~\cite{hou-etal-2019-identification}. 
Each article's parsed text was then finally annotated with (T, D, M, S) quadruples via distant labeling.

\noindent{\textbf{Corpus with no leaderboards.}} In addition to our base dataset reported in \autoref{table:dataset_stats}, we additionally included a set of approximately 4,401 and 548 articles that do not report leaderboards into the train and test sets, respectively. These articles were randomly selected by leveraging the arxiv category feature, then filtering it to papers belonging to domains unrelated to AI/ML/Stats. These articles were annotated with the \textit{unanswerable} label to finetune our language model in recognizing papers without (T,D,M,S) mentions in them.  

Our final corpus statistics are reported in \autoref{table:dataset_stats}. Since in this work, the model complexity and the time required to fine-tune a language model is far greater than the approaches we used in our previous work \cite{kabongo2023zero}, we only reported our experiments based on the results from Fold 1.  Furthermore, in the first main column, i.e. the ``Our corpus'' column, when compared with the corpus from existing work by \cite{hou-etal-2019-identification}, i.e. the ``Prior work'' column, our corpus shows itself to be significantly larger thus showing a more large-scale evaluation setting.
% This two-fold experimental setup is explained in detail later in \autoref{eval}.

\noindent{\textbf{The \textsc{SOTA} task objective.}} We phrased the following question to formulate our task objective w.r.t. the (T, D, M, S) extraction target: \textit{What are the values for the following properties to construct a Leaderboard for the model introduced in this article: task, dataset, metric, and score?} In essence, it encapsulates an IE task.

\noindent{\textbf{Instructions for the LLM.}} LLMs progress through initial pretraining and subsequent finetuning stages~\cite{unifiedqa,unifiedskg,supernaturalinstructions,p3,honovich2022unnatural,flan-collection}, but they might still struggle to interpret instructions. The practice of instruction finetuning~\cite{wei2021finetuned} has surfaced as an essential approach for augmenting the capability of LLMs to interpret and respond to instructions~\cite{lu-etal-2023-pivoine,wang2023instructuie}. As such the choice of the instruction is also crucial since it acts as a template that encodes the task and its objectives, instructing the LLM on how to achieve the specified objective.
%Instruction tuning~\cite{unifiedqa,unifiedskg,supernaturalinstructions,p3,honovich2022unnatural,flan-collection} enhances LLMs by providing explicit finetuning instructions, improving adaptability and performance across tasks and domains by offering specific guidance~\cite{instructgpt,flan-t5}, contrasting with traditional methods that use unlabeled data~\cite{t5,liu2019multi,muppet,ext5}. This approach streamlines finetuning and enables LLMs to execute tasks effectively with single instructions.

%Instruction tuning, as a novel approach~\cite{unifiedqa,unifiedskg,supernaturalinstructions,p3,honovich2022unnatural,flan-collection}, significantly enhances the performance of LLMs by providing explicit instructions during finetuning. These instructions guide the model's behavior~\cite{instructgpt,flan-t5}, making it more adaptable and effective in various learning scenarios. In contrast to traditional non-instruction tuning methods~\cite{t5,liu2019multi,muppet,ext5}, which rely solely on unlabeled data, instruction tuning introduces specific guidance, streamlining the finetuning process and improving performance on new tasks and domains~\cite{sanh2022multitask}. This approach enables the generic prompting of LLMs to perform different tasks with a single instruction, essentially serving as a template that encodes the task and its objectives, instructing the LLM on how to achieve the specified objective.

The ``Flan 2022 Collection'' is an extensive, open-source compilation of 62 previously released NLP datasets, organized into 12 task types including reading comprehension, sentiment analysis, natural language inference, and more, making it a vital resource for developing generic multi-task LLMs. Significantly, FLAN provided over \href{https://github.com/google-research/FLAN/blob/main/flan/templates.py}{10 human-curated natural instructions} per dataset, detailing the tasks, which we utilized to direct our LLM for complex IE tasks. We specifically chose instructions from the SQuAD\_v2~\cite{squad-v1,squad-v2} and DROP~\cite{drop} datasets, with 8 from SQuAD and 7 from DROP deemed appropriate. The general characteristic of the selected instructions, detailed in our appendix \ref{app:instructions}, is that they encode a context (in our case the DocTAET representation of an article) and the \textsc{SOTA} task objective, and instruct the model to fulfill the objective. %All instructions we used are elaborated in \autoref{app:instructions}. %These instructions, detailed in our appendix, are designed to encapsulate the context (using the DocTAET representation) and the task objective, guiding the LLM to achieve the specified goal.

%The ``Flan 2022 Collection'' was a large-scale open-sourced collection of 62 prior publicly released datasets in the NLP community clustered as 12 task types, such as reading comprehension (RC), sentiment, natural language inference (NLI), struct to text, etc. It is the most comprehensive resource facilitating open-sourced LLM development as generic multi-task models. Importantly, and of relevance to this work, FLAN was not just a super-amalgamation of datasets encapsulating different learning objectives, but also included at least \href{https://github.com/google-research/FLAN/blob/main/flan/templates.py}{10 human-curated natural instructions} per dataset that described the task for that dataset. As such, we select a set of instructions to guide the LLM for our complex IE task from the FLAN collection. Specifically, we identified the applicable instructions to our task were those designed for the SQuAD\_v2~\cite{squad-v1,squad-v2} and DROP~\cite{drop} datasets. Specifically, 8 SQuAD and 7 DROP instructions were found suitable. The general characteristic of the selected instructions is that they encode a context (in our case the DocTAET representation of an article) and the \textsc{SOTA} task objective, and instruct the model to fulfill the objective. All instructions we used are elaborated in \autoref{app:instructions}.

\section{Approach}
\label{model}

% of the instruction-tuned FLAN-T5 model~\cite{flan-t5}

%Our approach is single-task instruction-finetuning to address the \textsc{SOTA} task. As such it aims to be an incremental progression of the instruction-tuning paradigm introduced as FLAN (Finetuned Language Net)~\citep{flan,flan-t5,flan-collection}. Specifically \citep{flan-t5} ask: \textit{are instruction-finetuned models better for single-task finetuning?} as a recommendation for future work. Our work then is a direct examination of this research question except for a novel task type that the model is known to not have been explicitly trained for. 

Our approach examines the effectiveness of single-task instruction-finetuning on a novel task, i.e. the \textsc{SOTA} task, advancing the instruction-tuning paradigm initially proposed by FLAN (Finetuned Language Net)\citep{flan,flan-t5,flan-collection}. Equipped with the relevant set of 15 total instructions (8 SQuAD and 7 DROP), we needed to do two things: \textbf{1.} For each instance in the dataset, instantiate the ``Context'' placeholder in the instructions with the DocTAET context feature of a paper and the ``Question'' placeholder with formulated question for the SOTA objective. \textbf{2.} The LLM could then be finetuned with the instruction-instantiated training set. From \autoref{table:dataset_stats}, given our training dataset had approximately 7,987 (T, D, M, S) papers x 15 instructions x 1 SOTA objective question = 119,805 instruction-instantiated data points to train the LLM. To this, the 4,401 papers without leaderboards x 15 instructions x 1 SOTA objective = 66,015 instruction-instantiated data points were added. 

\subsection{Model}

We select the FLAN-T5 XL model~\cite{flan-t5} from its range of \href{https://github.com/google-research/t5x/blob/main/docs/models.md#flan-t5-checkpoints}{public checkpoints}, which come in various sizes (Small 80M, Base 250M, Large 780M, XL 3B, and XXL 11B). The choice of the Large model strikes a balance between the Small and XXL models, offering an ample number of parameters for our intricate IE task while remaining practical for deployment. This decision stems from considerations of efficiency, as extensive-scale LLMs were deemed impractical for a single task. Our choice of Flan-T5 was motivated by prior empiricism~\cite{flan-collection} proving instruction-tuned models as more computationally efficient starting checkpoints for new tasks -- FLAN-T5 required less finetuning to converge higher and faster than T5 on single downstream tasks~\citep{flan-collection}. Our model choice builds upon previous research, enhancing the T5 text-to-text sequence generation model~\citep{t5} with FLAN-T5~\citep{flan-t5} to improve alignment with instructions in unseen tasks and zero-shot settings. Our resulting model is called \textsc{SOTA}-Flan-T5.

\begin{table*}[!htb]
  \centering
  \begin{adjustbox}{width=1\textwidth}
  
    \begin{tabular}{|p{2cm}|p{1.1cm}cccc|ccccc|l|}
      % \hline
      % & & \multicolumn{5}{|c|}{Highest Scores} & \multicolumn{5}{c|}{Lowest Scores} \\
      \toprule
    \bf Instruction & \bf Rouge1 & \bf Rouge2 & \bf RougeL & \bf RougeLsum & \bf \stackbox[c]{General\\-Accuracy} &\
       \bf  Rouge1 & \bf Rouge2 & \bf RougeL & \bf RougeLsum & \bf \stackbox[c]{General\\-Accuracy} & \bf Instruction\\
      \midrule
    Drop 1 & 73/62 & 11/8 & 73/62 & 73/62 & 96/91 & 73/62 & 11/8 & 73/62 & 73/62 & 96/91 & Squad 1 \\ \hline
    Drop 2 & 73/62 & 11/8 & 73/62 & 73/62 & 96/91 & 72/62 & 11/8 & 72/63 & 72/62 & 95/91 & Squad 2 \\ \hline
    Drop 3 & 73/62 & 11/8 & 73/62 & 73/62 & 96/92 & 73/62 & 11/8 & 73/62 & 73/62 & 96/91 & Squad 3 \\ \hline
    Drop 4 & 73/62 & 11/8 & 73/62 & 73/62 & 96/91 & 73/62 & 11/8 & 73/62 & 73/62 & 96/91 & Squad 4 \\ \hline
    Drop 5 & 73/61 & 11/8 & 72/62 & 72/61 & 96/91 & 73/62 & 11/8 & 73/62 & 73/62 & 96/91 & Squad 5 \\ \hline
    Drop 6 & 73/62 & 11/8 & 73/62 & 73/62 & 96/91 & 73/62 & 11/8 & 73/62 & 73/62 & 96/91 & Squad 6 \\ \hline
    Drop 7 & 73/61 & 11/8 & 73/61 & 73/61 & 96/90 & 73/63 & 11/8 & 73/63 & 73/63 & 96/92 & Squad 7 \\ \hline
    -      & -   & -  & -  & -  & -   & 73/62 & 11/8 & 73/62 & 73/62 & 96/91 & Squad 8 \\ \hline
    \end{tabular}
  \end{adjustbox}
  \caption{Evaluation results of SOTA-Flan-T5 Large with output evaluations as a structured summary generation task (reported with ROUGE metrics) as well as binary classification between papers with and without leaderboards (reported as General Accuracy) for each of the 15 instructions from DROP and SQuAD datasets vs w/o templates instruction, respectively.}
  \label{tab:merged-results-rouge}
\end{table*}

\begin{table*}[!htb]
\centering
\begin{adjustbox}{width=1\textwidth}
    \begin{tabular}{|l|l|ccccc|ccccc|l|} \hline
        &         & \multicolumn{5}{|c|}{Drop Instructions} & \multicolumn{5}{c|}{SQuAD v2 Instructions} &     \\ \hline
        &  & \bf Task  & \bf Dataset & \bf Metric & \bf Score & \bf Overall & \bf Task & \bf Dataset & \bf Metric & \bf Score & \bf Overall & \\ \hline
        \multirow{2}{*}{D1} &  \textcolor{orange}{Exact} & 36/14 & 12/08 & 24/12 & 0.2/0.1 & 18/08 & 37/14 & 13/08 & 24/12 & 0.1/0.2 & 18/09 & \multirow{2}{*}{S1} \\ \cline{2-12}
        & \textcolor{blue}{Partial} & 55/28 & 22/17 & 36/18 & 0.2/0.4 & 28/16 & 55/29 & 23/18 & 37/17 & 0.1/0.3 & 29/16 & \\ \hline
        \multirow{2}{*}{D2} &  \textcolor{orange}{Exact} & 36/14 & 12/08 & 23/12 & 0.1/0.2 & 18/09 & 35/14 & 12/08 & 22/13 & 0.1/0.2 & 17/09 & \multirow{2}{*}{S2} \\ \cline{2-12}
        & \textcolor{blue}{Partial} & 55/29 & 23/18 & 36/17 & 0.1/0.3 & 28/16 & 54/29 & 21/18 & 35/18 & 0.1/0.4 & 27/16 & \\ \hline
        \multirow{2}{*}{D3} &  \textcolor{orange}{Exact} & 36/14 & 12/08 & 23/12 & 0.1/0.2 & 18/09 & 36/14 & 12/08 & 23/12 & 0.1/0.2 & 18/09 & \multirow{2}{*}{S3} \\ \cline{2-12}
        & \textcolor{blue}{Partial} & 55/29 & 23/18 & 36/17 & 0.1/0.3 & 29/16 & 37/29 & 12/18 & 23/17 & 0.1/0.5 & 18/16 & \\ \hline
        \multirow{2}{*}{D4} &  \textcolor{orange}{Exact} & 36/14 & 13/08 & 23/12 & 0.1/0.2 & 18/09 & 37/14 & 12/08 & 23/12 & 0.1/0.2 & 18/09 & \multirow{2}{*}{S4} \\ \cline{2-12}
        & \textcolor{blue}{Partial} & 55/29 & 23/18 & 36/18 & 0.1/0.5 & 29/16 & 55/29 & 23/18 & 36/17 & 0.1/0.5 & 29/16 & \\ \hline
        \multirow{2}{*}{D5} &  \textcolor{orange}{Exact} & 36/14 & 13/08 & 25/12 & 0.1/0.2 & 18/08 & 37/14 & 12/08 & 23/12 & 0.1/0.2 & 18/09 & \multirow{2}{*}{S5} \\ \cline{2-12}
        & \textcolor{blue}{Partial} & 56/29 & 22/17 & 37/18 & 0.1/0.3 & 29/16 & 55/28 & 23/18 & 36/17 & 0.1/0.5 & 29/16 & \\ \hline
        \multirow{2}{*}{D6} &  \textcolor{orange}{Exact} & 36/14 & 12/08 & 23/12 & 0.1/0.2 & 18/09 & 37/14 & 12/08 & 23/12 & 0.1/0.2 & 18/09 & \multirow{2}{*}{S6} \\ \cline{2-12}
        & \textcolor{blue}{Partial} & 55/29 & 23/18 & 36/18 & 0.1/0.5 & 29/16 & 55/29 & 23/17 & 36/17 & 0.1/0.5 & 29/16 & \\ \hline
        \multirow{2}{*}{D7} &  \textcolor{orange}{Exact} & 36/14 & 13/8 & 24/12 & 0.1/0.2 & 18/09 & 35/14 & 12/08 & 23/12 & 0.1/0.2 & 18/09 & \multirow{2}{*}{S7} \\ \cline{2-12}
        & \textcolor{blue}{Partial} & 56/28 & 22/17 & 36/17 & 0.1/0.5 & 29/16 & 56/29 & 22/18 & 35/18 & 0.1/0.3 & 28/16 & \\ \hline
        \multirow{2}{*}{-} &  \textcolor{orange}{Exact} & - & - & - & - & - & 35/14 & 12/08 & 23/12 & 0.1/0.2 & 18/09 & \multirow{2}{*}{S8} \\ \cline{2-12}
        & \textcolor{blue}{Partial} & - & - & - & - & - & 55/29 & 22/18 & 35/18 & 0.1/0.5 & 28/16 & \\ \hline
    \end{tabular}
\end{adjustbox}
\caption{Evaluation results of SOTA-Flan-T5 Large w.r.t. the individual (Task, Dataset, Metric, Score) elements and Overall in the model JSON generated output in terms of F1 score for each of the 15 instructions from DROP and SQuAD datasets vs w/o templates instruction respectively.
}
\label{tab:merged-results-f1}
\end{table*}

\section{Evaluations}
\label{eval}

\noindent{\textbf{Experimental setup.}} For training, we had one main experimental setting based on the 15 instructions. As elicited earlier in \autoref{sec:corpus}, i.e. the Corpus section, each of the 15 instruction were instantiated with the 12,388 (T, D, M, S) data instances including both papers with leaderboard and w/o leaderboards and the SOTA question resulting in a total of 185,820 instances to instruction finetune Flan-T5 Large. In this scenario, we hypothesized that this repetition in the data instances across the instructions would cause the resulting model to overfit the training dataset. Thus to control for this, we applied the following experimental setup. Each instruction was instantiated with a random selection of only half the total templates occurrences of every data instances resulting in a finetuning dataset of a sizeable 92,910 instances. In the test scenario, however, we report per instruction (T, D, M, S) instantiated data results. As shown in \autoref{table:dataset_stats}, for the test set with approximately 241 (T, D, M, S) and 548 papers with and without leaderboards respectively, evaluations results are shown for each instruction separately with a total of 789 underlying papers representing those with and without leaderboards. Model hyperparamter details are in \autoref{app:hyp}. In terms of compute, all experiments including inference were run on an NVIDIA h100 GPU. 
% Training took ~50 hours on the 50\% sampled dataset, while inference lasted ~10 minutes for 2,353 test instances.

\noindent{\textbf{Metrics.}} We evaluated the SOTA-Flan-T5 model in two settings. In the first setting, we treated the \textsc{SOTA} task objective as a structured summarization task. In this setting, we applied standard summarization ROUGE metrics~\cite{rouge} (details in \autoref{app:rouge}). Furthermore, we also tested the models ability to identify papers with leaderboards and those without. This task was simple. For the papers with leaderboards, the model replied with a structured summary and for those it identified as without it replied as ``unanswerable.'' For these evaluations we applied simple accuracy measure. In the second setting, we evaluated the model JSON output in a fine-grained manner w.r.t. each of the individual (T, D, M, S) elements and overall for which we reported the results in terms of the standard F1 score.

\subsection{Results and Discussion}

\noindent{\textbf{Structured summary generation evaluations.}} \autoref{tab:merged-results-rouge} results show model's capacity in generating structured summaries per the \textsc{SOTA} objective. The results obtained were consistent across all 15 instructions which indicates that the model systematically follows all instructions and handles them all in more or less the same way. Notably, the general accuracy, i.e. the ability of the model to discriminate between papers with leaderboards and those without is nearly perfect at 95\% indicating a core strength of the model. 
% As it can be seen from table \autoref{tab:rouge-no-templates} the use of FLAN-T5 instructions collections has improved our model by a large margin .

The ROUGE metrics, which measure the overlap between the model's output and reference summaries, have improved by approximately 10 points for ROUGE1 and 3 points for ROUGE2 when comparing the instruction-conditioned model to the baseline shown in \autoref{tab:merged-results-f1}. The improvement is indicative of the model's enhanced ability to generate summaries that are not only more aligned with human judgments but also more informative and concise.

\noindent{\textbf{\textsc{SOTA} objective (Task, Dataset, Metric, Score) element-wise generation evaluations.}} Next we examine the results reported in \autoref{tab:merged-results-f1}. Specifically, we examine how well the finetuned \textsc{SOTA}-Flan-T5 model performs when evaluated to precisely extract each of the \textsc{SOTA} objective elements i.e. the Task, Dataset, Metric, and Score in a response produced as one or more related quadruples per paper. These results are reported in terms of F1 scores in an exact match and partial match settings of the model output. Consistent with the results in \autoref{tab:merged-results-rouge}, the model responds consistently across the DROP and SQuAD instruction types. Understandably the results in the exact match setting are at least 10 points lower than the results in the partial match setting. We see that across all four elements, the Task is easiest to extract at $\sim$36\% exact-match evaluations and $\sim$56\% partial-match evaluations. The Metric element was shown to be second easiest to extract at $\sim$25\% exact-match and $\sim$37\% partial-match evaluations followed the Dataset element at $\sim$13\% exact-match and $\sim$23\% partial-match evaluations. The model failed in extracting the Score element indicating that an alternate strategy is warranted here. Conclusively, we started out with the research question that examined whether \textsc{SOTA} addressed in a task generation objective would work at all and whether the resulting LLM would be effective? Examining the results in the ``Overall'' column we see our approach is competitive with the prior state-of-the-art, i.e. the AxCell system~\cite{axcell}. Additionally, a point to note here is that our labels are all unnormalized and obtained directly from the community-curated PwC labels which can account in part for lower scores by our approach and our zeroshot test set contains at least a leaderboard that was not seen at training time. In this case, our annotated test dataset with distantly supervised (Task, Dataset, Metric, Score) annotations versus our LLM predictions can be browsed here \url{https://scinext-project.github.io/#/sota}.

The incorporation of the FLAN-T5 instruction collection into our model's training regimen has demonstrably enhanced its performance across both structured summarization and SOTA Objective tasks. This effect is quantitatively evident in the results presented in \autoref{tab:merged-results-rouge} and \autoref{tab:merged-results-f1}, which showcases a consistent improvement in ROUGE scores as well as Task, Dataset, Metric and Score element-wise F1 Score when the model is conditioned with FLAN-T5 instructions.

% Finally, we also evaluated models trained in all the instances as opposed to a random selection of 50\%. The resulting model scores were not improved. These results can be found in \autoref{app:res}.

\section{Error Analysis}

In this section, we perform the error analysis of our finetuned \textsc{SOTA}-Flan-T5 model. 

\noindent{\textbf{Type 1 - Missing information}} The most prominent cause of error in the leaderboard generation is the need for the appropriate entities of interest in the provided context, which refers to our DOCTEAT in this context. FLAN-T5 family of models suffers from the same limitation of 512 max token length caused by the quadratic nature of the underlying attention mechanism. Similarly to \cite{hou-etal-2019-identification}, we obtained a summarized version of the paper, called \textbf{DOCTEAT} which stands for \textbf{DOC}ument, \textbf{T}itle, \textbf{A}bstract, \textbf{E}xpSetup, and \textbf{T}ableInfo. We noticed that, the abstracted representation doesn't usually contain the score numeric value associated with the dataset and metric reported in a paper. Thus making the Language Model learn to generate a value that is not available in the context.

\noindent{\textbf{Type 2 - Crowdsourced label discrepancies }} Discrepancies between all the Tasks, Datasets, Metrics, and Scores reported in a particular paper vs the metadata available from papers with code data dump is a cause of confusion in the LLM training. We noticed instances of papers with code leaderboard mentions unrelated to the paper's main contribution, and cases of mentions completely unrelated to the paper. We noticed that, the language model tend to learn the grammatical structure leading to the mention of these entities in the DOCTEAT, but struggle to learn the appropriate representation caused by the misalignment between the leaderboard entities captured by papers with code compared to the ground truth leaderboard addressed in the paper. Thus, an extra human validation of the dataset curated through PWC becomes necessary for future experiments.

\section{Conclusions and Future Work}

In this paper, we have demonstrated how LLMs can be leveraged for the automated construction of leaderboards from empirical AI research papers modeled as the \textsc{SOTA} objective. As such, we specifically investigated instruction finetuning of the FLAN-T5 model. Our experimental results showed that the finetuned SOTA-Flan-T5 model was effective for the task. This in turn impacts future directions for the task from an NLI paradigm aptly situating it in the area of LLM research as a text generation paradigm instead.

\section{Limitations}
Our approach depends heavily on the quality of data processing and the inherent limitations of the tools employed, such as Pandoc, for converting LaTeX documents to plain text. Errors introduced during this conversion can significantly affect the extraction accuracy of (Task, Dataset, Metric, Score) quadruples. Additionally, our model's generalizability across various domains of academic research beyond computer science is not yet verified. The distinct formats and terminologies prevalent in different disciplines may pose a challenge, and as such, the model's applicability across these varied fields remains a topic for future research.

\section{Ethics Statement}
The datasets used in this study were sourced from the arXiv repository, adhering to open access policies. Despite this, the automated nature of our information extraction poses ethical considerations, primarily due to potential misinterpretations or oversimplifications of nuanced academic content. The potential of propagation errors from source materials to final outputs due to preprocessing tools underscores the need for clear communication regarding these limitations to users of our system. This is crucial to ensure that the information provided through the generated leaderboards accurately reflects the advancements in AI research without misleading the academic community or the public.

\bibliography{main}

% \bibliography{custom}

\appendix

\section{Instructions: Qualitative Examples}
\label{app:instructions}

In this section, we elicit each of the instructions that were considered in this work as formulated in the FLAN 2022 Collection for the SQuAD\_v2 and DROP datasets.

\subsection{The Stanford Question Answering Dataset (SQuAD\_v2)}

\paragraph{Instruction 1 (S1):}
\mbox{}\\
\{Context\} \textbackslash n\textbackslash n Please answer a question about this article. If the question is unanswerable, say "unanswerable". \{Question\}

\paragraph{Instruction 2 (S2):}
\mbox{}\\
\{Context\} \textbackslash n \{Question\} If the question is unanswerable, say "unanswerable"

\paragraph{Instruction 3 (S3):}
\mbox{}\\
\{Context\}\textbackslash n Try to answer this question if possible (otherwise reply "unanswerable"): \{Question\}

\paragraph{Instruction 4 (S4):}
\mbox{}\\
\{Context\} \textbackslash n\textbackslash n Please answer a question about this article. If the question is unanswerable, say "unanswerable". \{Question\}'
\{Context\}  \textbackslash n Try to answer this question if possible (otherwise reply "unanswerable"): \{Question\}

\paragraph{Instruction 5 (S5):}
\mbox{}\\
\{Context\}\textbackslash n If it is possible to answer this question, answer it for me (else, reply "unanswerable"): \{Question\}

\paragraph{Instruction 6 (S6):}
\mbox{}\\
\{Context\}\textbackslash n \textbackslash n Answer this question, if possible (if impossible, reply "unanswerable"): \{Question\}

\paragraph{Instruction 7 (S7):}
\mbox{}\\
Read this: \{Context\}\textbackslash n \textbackslash n \{Question\} \textbackslash n What is the answer? (If it cannot be answered, return "unanswerable")

\paragraph{Instruction 8 (S8):}
\mbox{}\\
Read this: \{Context\}\textbackslash n Now answer this question, if there is an answer (If it cannot be answered, return "unanswerable"): \{Question\}

\subsection{Discrete Reasoning over Paragraphs (DROP) Dataset}

\paragraph{Instruction 1 (D1):}
\mbox{}\\
Answer based on context:\textbackslash n \textbackslash n \{Context\}\textbackslash n \textbackslash n \{Question\}

\paragraph{Instruction 2 (D2):}
\mbox{}\\
\{Context\}\textbackslash n \textbackslash n Answer this question based on the article: \{Question\}

\paragraph{Instruction 3 (D3):}
\mbox{}\\
\{Context\}\textbackslash n \textbackslash n \{Question\}

\paragraph{Instruction 4 (D4):}
\mbox{}\\
\{Context\}\textbackslash n Answer this question: \{Question\}

\paragraph{Instruction 5 (D5):}
\mbox{}\\
Read this article and answer this question \{Context\}\textbackslash n \{Question\}

\paragraph{Instruction 6 (D6):}
\mbox{}\\
\{Context\}\textbackslash n \textbackslash n Based on the above article, answer a question. \{Question\}

\paragraph{Instruction 7 (D7):}
\mbox{}\\
Context: \{Context\}\textbackslash n \textbackslash n Question: \{Question\}\textbackslash n \textbackslash n Answer:

\section{Our Experimental Hyperparamters}
\label{app:hyp}
We used two main experimental settings in this work. The first consists of a dataset of a randomly selected half of every individual template instance, and the second one is a dataset with no template instances called baseline in the paper.

Given that the average context length of our dataset was close to the 512 sequence length limit by T5 and the size of the available GPU, a batch size of 4 and gradient\_accumulation\_steps of 1 were used. All experiments were run on five epochs and we used AdafactorSchedule and Adafactor optimizer~\cite{shazeer2018adafactor} with scale\_parameter=True, relative\_step=True, warmup\_init=True, lr=None.

The evaluations were all done on a dataset made of individual template instructions separately, as reported in table \autoref{tab:merged-results-rouge}. %to \autoref{tab:f1-all}.

% The evaluations were done on each epoch on the dev set and we kept two best (the one maximizing the "Overall Partial F1" score) and last checkpoints in each model training process to then use for inference on test set.

\section{ROUGE Evaluation Metrics}
\label{app:rouge}

The ROUGE metrics~\cite{rouge} are commonly used for evaluating the quality of text summarization systems. ROUGE-1 measures the overlap of unigram (single word) units between the generated summary and the reference summary. ROUGE-2 extends this to measure the overlap of bigram (two consecutive word) units. ROUGE-L calculates the longest common subsequence between the generated and reference summaries, which takes into account the order of words. ROUGE-LSum is an extension of ROUGE-L that considers multiple reference summaries by treating them as a single summary. These metrics provide a quantitative assessment of the similarity between the generated and reference summaries, helping researchers and developers evaluate and compare the effectiveness of different summarization approaches. They have become widely used benchmarks in the field of automatic summarization.

\end{document}